\pdfoutput=1

\documentclass[11pt]{article}

\usepackage[preprint]{acl}

\usepackage{times}
\usepackage{latexsym}

\usepackage[T1]{fontenc}

\usepackage[utf8]{inputenc}

\usepackage{microtype}

\usepackage{inconsolata}

\usepackage{graphicx}

\title{An Interdisciplinary Review of Commonsense Reasoning and Intent Detection}

\author{Md Nazmus Sakib \\
  University of Maryland, Baltimore County \\
  \texttt{msakib1@umbc.edu} 
  }

\begin{document}
\maketitle
\begin{abstract}
This review explores recent advances in commonsense reasoning and intent detection, two key challenges in natural language understanding. We analyze 28 papers from ACL, EMNLP, and CHI (2020–2025), organizing them by methodology and application. Commonsense reasoning is reviewed across zero-shot learning, cultural adaptation, structured evaluation, and interactive contexts. Intent detection is examined through open-set models, generative formulations, clustering, and human-centered systems. By bridging insights from NLP and HCI, we highlight emerging trends toward more adaptive, multilingual, and context-aware models, and identify key gaps in grounding, generalization, and benchmark design.
\end{abstract}

\section{Introduction}

Commonsense reasoning is an important problem in natural language understanding. It helps models make inferences that match human knowledge about the world, such as cause and effect, physical actions, social behavior, and hidden assumptions \cite{sap2019socialiqa, aggarwal2021explanations}. Unlike models that only rely on surface-level text, commonsense reasoning allows systems to fill in missing information, reduce ambiguity, and understand what is not directly said. This ability is especially important in dialogue systems, where understanding the speaker's intent often depends on shared background knowledge \cite{wu2020diverse}. 

Intent detection is the task of identifying the purpose behind a speaker's utterance. It is usually treated as a classification problem with a fixed set of intent labels \cite{mensio2018multi}. However, in real-world settings, people often express their intentions in indirect ways. To handle this, models need to rely on pragmatic and commonsense understanding \cite{pareti2013automatically, louis2020d}. For example, knowing whether a question shows confusion, politeness, or sarcasm may require the system to infer social and contextual cues \cite{rashkin2018towards}.

Recently, several studies have explored how commonsense knowledge can support intent detection and related tasks. These studies use different sources of knowledge, such as structured graphs \cite{lin2019kagnet}, pretrained language models \cite{bosselut2019comet}, and knowledge generated through prompting \cite{liu2021generated}. At the same time, researchers have developed new benchmarks that test models' reasoning skills beyond surface similarity, focusing on causal reasoning, generalization, and understanding implicit meaning \cite{tafjord2019quartz, chi2024unveiling}.

This review gives an updated overview of recent work on commonsense reasoning and intent detection, focusing on papers published between 2020 and 2025 in ACL, EMNLP, and CHI. Unlike earlier surveys, we include a wider range of methods, such as graph-based, generative, contrastive, and hybrid models, and cover different types of reasoning like causal, physical, and social. We also explore how these methods are used in intent detection, especially in multilingual, interactive, and user-centered settings. By combining work from both NLP and HCI fields, this review offers a more interdisciplinary view and points out important directions for future research that connects reasoning with intent understanding in real-world applications.

\section{Related Works}
Prior surveys in commonsense reasoning and intent detection have primarily focused on knowledge representation frameworks or specific downstream tasks such as question answering or dialogue generation. For instance, \citet{yu2024natural} provide a comprehensive overview of natural language reasoning in NLP. and the integration of commonsense knowledge into NLP tasks. In the area of conversational AI, \citet{richardson2023commonsense} examine the role of commonsense reasoning in dialogue systems, discussing relevant datasets and approaches. Other reviews have assessed commonsense reasoning from a systems perspective, outlining how benchmarks are structured and revealing inconsistencies in how such knowledge is evaluated \cite{losey2018review}. Meanwhile, \citet{weld2022survey} emphasize the integration of physical, social, and intuitive commonsense into NLP and LLM-driven systems.

Regarding intent detection, \citet{weld2022survey} analyze trends and challenges in joint models for intent classification and slot filling. Additionally, \citet{jbene2025intent} evaluate the performance of different neural network architectures in intent recognition. Broader reviews further extend this perspective by considering intent inference in physical human–AI interaction, where real-time prediction of user intent is critical for shared control and feedback \cite{liu2019review, davis2015commonsense}. Joint modeling of intent and slots has also been critically evaluated, particularly in regard to system performance under multi-task constraints \cite{sap2020commonsense}.

Previous surveys on commonsense reasoning and intent detection have mostly focused on knowledge representation or specific applications like question answering and dialogue systems. They rarely integrate diverse modeling approaches or consider human-centered use cases. Moreover, they are largely confined to NLP venues such as ACL and EMNLP, often overlooking insights from HCI forums like CHI. Our review expands on this by covering a broader set of papers from ACL, EMNLP, and CHI (2020–2025), organized into themes such as zero-shot modeling, cultural adaptation, structured reasoning, and user-focused applications. By reviewing both NLP and HCI work, we offer a more interdisciplinary and updated perspective on reasoning and intent understanding, including multilingual evaluation, generative intent labeling, and interactive systems.

\section{Methodology}
We conducted a structured review of commonsense reasoning methods in NLP with a focus on their relevance to intent detection and dialogue understanding. The keywords used in our search were selected based on terminology frequently appearing in prior surveys, benchmark datasets, and recent ACL, EMNLP, and CHI publications. Terms like "commonsense reasoning," "intent detection," and "social inference" were included, along with others related to methodology and context. The full list is in Appendix \ref{appendix}. We focused on peer-reviewed papers from top conferences between 2020 and 2025 (ACL, EMNLP, CHI). We excluded preprints and unrelated downstream tasks for reviewing. After initial screening, 28 papers were selected for detailed review. These were grouped into categories based on methodological and application-based approach (graph-based, generative, prompting or hybrid) and reasoning type (causal, dialogic, social). The aim of this review is to synthesize key research trends, critically compare existing methodologies, and highlight ongoing challenges at the intersection of commonsense reasoning and intent detection in natural language processing. 

\section{Result}
We have divided our selected papers into two main themes and then into subthemes. The following part of this section will discuss about each subthemes in details. The categorization of all the papers is shown in Table \ref{tab:categorization}.

\begin{table*}[ht]
\centering
\begin{tabular}{|p{3.5cm}|p{5.5cm}|p{6cm}|}
\hline
\textbf{Main Category} & \textbf{Sub-category} & \textbf{Citations} \\
\hline
Commonsense Reasoning & Self-supervised and Zero-shot Learning & \cite{shwartz-etal-2020-unsupervised}; \cite{klein-nabi-2021-towards}; \cite{lin-etal-2021-differentiable}; \cite{murata2024time} \\
\hline
Commonsense Reasoning & Multilingual and Cultural Adaptation & \cite{lin-etal-2021-common}; \cite{yin-etal-2021-broaden}; \cite{sakai-etal-2024-mcsqa} \\
\hline
Commonsense Reasoning & Structured Reasoning and Evaluation Analysis & \cite{saha-etal-2021-explagraphs}; \cite{branco-etal-2021-shortcutted}; \cite{hwangyatomic2020}; \cite{he-etal-2020-box} \\
\hline
Commonsense Reasoning & Interactive, Dialog-based, and Applied Commonsense & \cite{ghosal-etal-2022-cicero}; \cite{romero-razniewski-2022-children}; \cite{chen-etal-2023-say}; \cite{qu-etal-2022-commonsense}; \cite{xu2023engage} \\
\hline
Intent Detection & Open-set and Zero-shot Detection & \cite{yan-etal-2020-unknown}; \cite{ouyang-etal-2021-energy}; \cite{choi-etal-2021-outflip}; \cite{wu-etal-2021-label} \\
\hline
Intent Detection & Multi-intent Modeling and Generative Formulation & \cite{qin-etal-2020-agif}; \cite{zhang-etal-2024-discrimination} \\
\hline
Intent Detection & Contrastive Learning and Clustering & \cite{kumar-etal-2022-intent}; \cite{zhang-etal-2022-new} \\
\hline
Intent Detection & Human-centered and HCI Applications & \cite{csencan2024intention}; \cite{yu2023history}; \cite{belardinelli2024gaze}; \cite{reese2024comparing} \\
\hline
\end{tabular}
\caption{Categorization of reviewed papers by main theme and sub-theme}
\label{tab:categorization}
\end{table*}

\subsection{Commonsense Reasoning}
We have categorized all the papers related to commonsense reasoning to four sub-themes: (1) Self-supervised and Zero-shot Learning, (2) Multilingual and Cultural Adaptation, (3) Structured Reasoning and Evaluation Analysis, and (4) Interactive, Dialog-based, and Applied Commonsense.

\subsubsection{Self-supervised and Zero-shot Learning}
Modern approaches often move beyond supervised training, relying on internal model dynamics or indirect supervision. The "self-talk" approach allows a model to pose and answer internal queries, improving zero-shot QA by enhancing internal knowledge activation \cite{shwartz-etal-2020-unsupervised}. Similarly, \citet{klein-nabi-2021-towards} applied perturbation-based refinement to improve Winograd Schema Challenge performance, achieving competitive results without external supervision. \citet{lin-etal-2021-differentiable} further extended this trend by proposing DrFact, a differentiable open-ended reasoning system using multi-hop retrieval from commonsense corpora. The system shows strong improvements but still relies on high-quality corpora, which may not cover all needed knowledge. Another work introduces COMET, a model that uses transformers to generate commonsense knowledge graphs, building upon existing resources like ConceptNet \cite{murata2024time}. However, the generated knowledge can be noisy sometimes and lacks grounding in specific contexts. 

\subsubsection{Multilingual and Cultural Adaptation}
Commonsense knowledge is often culturally embedded. \citet{lin-etal-2021-common} addressed this limitation with the Mickey Corpus and multilingual pretraining to improve non-English commonsense QA. Despite success, the model risks inheriting English-centric biases through translated datasets. \citet{yin-etal-2021-broaden} explored cultural bias in visual reasoning. Their GD-VCR dataset reveals that models trained on Western imagery perform poorly on scenes from Africa and Asia, especially when reasoning requires cultural understanding (e.g., ceremonies). Another multilingual dataset named X-CSQA, is designed to evaluate and improve commonsense reasoning across different languages \cite{sakai-etal-2024-mcsqa}. However, many translated questions preserve English-centric logic and cultural assumptions.

\subsubsection{Structured Reasoning and Evaluation Analysis}
To evaluate structured reasoning, \citet{saha-etal-2021-explagraphs} introduced ExplaGraphs, where models generate explanation graphs to justify stance classification. Results show that current systems fall short of human explanation quality. \citet{branco-etal-2021-shortcutted} raised fundamental concerns about the reliability of commonsense benchmarks, showing that models exploit shallow dataset artifacts instead of reasoning. Another paper presents ATOMIC 2020, a comprehensive commonsense knowledge graph that enhances the reasoning capabilities of AI models \cite{hwangyatomic2020}. Though the symbolic structure offers high coverage, it suffers from sparsity. \citet{he-etal-2020-box} extended this idea into NMT, revealing that translation systems struggle with disambiguation that requires implicit knowledge. 

\subsubsection{Interactive, Dialog-based, and Applied Commonsense}
Commonsense reasoning is often embedded in social interaction and social computing area. \citet{ghosal-etal-2022-cicero} introduced CICERO, a large annotated dialogue corpus where utterances are tagged with motivations, reactions, and other inferences. Such dialogue-specific reasoning challenges the generalizability across domains. \citet{romero-razniewski-2022-children} suggested training models with children's books to teach simple, explicit commonsense. Their childBERT system performs well, indicating that less abstract texts may benefit model training. However, such texts cover mostly basic knowledge and exclude nuanced or adult-level reasoning.

\citet{chen-etal-2023-say} examined how large language models (LLMs) fail at generating negative commonsense (e.g., "Lions don't live in the ocean"), even when they can handle the equivalent question-answer form. This shows a structural bias in generative training objectives. \citet{qu-etal-2022-commonsense} applied commonsense reasoning to e-commerce by creating a benchmark for "salient" commonsense facts that are relevant to product entities. Their work moves toward task-specific commonsense selection rather than general retrieval. And, \citet{xu2023engage} proposed an HCI approach where children co-participate with AI in storytelling dialogues to train systems in commonsense reasoning. The system is experimental, but reflects a broader vision of collaborative knowledge learning.

\subsection{Intent Detection}
For the intent detection, we have divided the papers in multiple directions: (1) Open-set and Zero-shot Detection, (2) Multi-intent Modeling and Generative Formulation, (3) Contrastive Learning and Intent Discovery, (4) Human-centered and HCI Applications.

\subsubsection{Open-set and Zero-shot Detection}
\citet{yan-etal-2020-unknown} used a semantic Gaussian mixture model (SEG) to identify out-of-distribution utterances based on learned intent clusters, offering a robust open-set detection framework. \citet{ouyang-etal-2021-energy} used energy-based modeling and synthetically generated "pseudo-OOD" utterances to boost detection ability. \citet{choi-etal-2021-outflip} offered a complementary adversarial strategy via HotFlip attacks that perturb known intent utterances to simulate unseen cases. \citet{wu-etal-2021-label} framed zero-shot multi-intent classification through label-aware embeddings. Their LABAN model projects utterances into a shared space with intent labels, achieving high performance on unseen intent types.

\subsubsection{Multi-intent Modeling and Generative Formulation}
To handle utterances with more than one intent, \citet{qin-etal-2020-agif} proposed AGIF, a graph-based model that dynamically integrates slot and intent features, showing strong performance on both multi- and single-intent datasets. Meanwhile, \citet{zhang-etal-2024-discrimination} reframed intent detection as a generative task. Their Gen-PINT system uses prompts to generate intent labels from utterances in low-resource or few-shot setups.

\subsubsection{Contrastive Learning and Clustering}
Discovery of unseen intents without full supervision is also studied. \citet{kumar-etal-2022-intent} applied deep contrastive clustering on partially labeled user logs, enabling dynamic intent categorization. \citet{zhang-etal-2022-new} enhanced this idea with a two-stage strategy: pre-training on known data and fine-tuning with contrastive loss to better isolate novel intents. Both works show that clustering-based approaches can reduce annotation cost, though they depend on accurate semantic separation.

\subsubsection{Human-centered and HCI Applications}
Real-world HCI contexts create specialized needs for intent detection. \citet{csencan2024intention} developed a classifier to detect self-harm intent from search queries, supporting mental health interventions. Their work highlights how intent modeling must balance sensitivity and specificity in critical use cases. \citet{yu2023history} designed a voice assistant for older adults, focusing on UI-navigation queries like "where is history?". Their study identifies key intent types and challenges in parsing help-seeking utterances. \citet{belardinelli2024gaze} provided a broader survey on gaze-based intent estimation, which uses visual attention transformed into text to infer user goals in UI interactions, although this line is still mostly exploratory. \citet{reese2024comparing} compared ChatGPT to human subjects in the Japanese Winograd Schema Challenge, noting that the model underperforms in linguistically subtle and culturally specific reasoning. This study links with commonsense reasoning and emphasizes cross-linguistic limitations of current LLMs.

\section{Discussion}

This review reveals a methodological shift in commonsense reasoning and intent detection from supervised learning toward more adaptive, context-aware approaches. Zero-shot and generative methods exemplify efforts to reduce reliance on labeled data while expanding generalization capacity. However, challenges remain around grounding: systems like COMET and DrFact extend knowledge bases and retrieval capabilities, but often lack contextual anchoring, limiting their practical reasoning depth. Cultural and multilingual adaptation has gained attention, yet many resources still exhibit English-centric biases. Though datasets like X-CSQA and Mickey aim to diversify language coverage, translated content frequently retains Western logic, undermining efforts toward true cross-cultural reasoning. This tension is also present in visual and symbolic reasoning systems, where knowledge sparsity and cultural abstraction persist despite broader coverage. Structured reasoning frameworks improve interpretability, but findings from works like ExplaGraphs and \citet{branco-etal-2021-shortcutted} caution against overreliance on benchmark performance, which may mask shallow heuristics. Interactive and dialog-based applications reflect a promising shift: Systems trained in pedagogical or social dialogue corpora foreground situational common sense and user alignment, although domain limitations remain.

In intent detection, open-set and contrastive models show progress in handling unseen or overlapping intents. Still, their success depends heavily on embedding quality and semantic separation. Generative intent labeling provides flexibility in low-resource settings, but introduces challenges of consistency and evaluation. Clustering based methods reduce annotation cost but risk semantic noise. Finally, HCI-focused work marks an essential evolution in framing intent not just as a classification task but as a design problem. Systems that address vulnerable users, such as older adults or individuals in distress, highlight the importance of interpretability, fairness, and contextual awareness. These findings point to a broader convergence: future systems must balance robustness with cultural sensitivity, task specificity with generalization with human-centered values.

\bibliography{main}

\appendix
\section{Keywords Used to Look For Papers}
\label{appendix}
"commonsense reasoning" "commonsense inference" "intent detection" "dialogue intent" "pragmatic inference" "context-aware intent classification" "commonsense in dialogue" "situated reasoning", "knowledge graph" "commonsense knowledge" "conceptnet" "social commonsense" "causal reasoning" "multi-hop reasoning" "language model prompting" "neural-symbolic reasoning" "zero-shot intent detection" "few-shot commonsense reasoning" "commonsense benchmarks", "social inference"

\end{document}